# A RESNET-BASED UNIVERSAL METHOD FOR SPECKLE REDUCTION IN OPTICAL COHERENCE TOMOGRAPHY IMAGES

*Ning Cai[1,3], Fei Shi[4], Dianlin Hu[1,3], Yang Chen[1,2,3]*

[1]Laboratory of Image Science and Technology, School of Computer Science and Engineering, Southeast University, Nanjing, China
[2]Centre de Rechercheen Information Biomedicale Sino-Francais (LIA CRIBs), Rennes, France
[3]The Key Laboratory of Computer Network and Information Integration (Southeast University), Ministry of Education, China
[4] School of Electronics and Information Engineering, Soochow University, China

## ABSTRACT

In this work we propose a ResNet-based universal method for speckle reduction in optical coherence tomography (OCT) images. The proposed model contains 3 main modules: Convolution-BN-ReLU, Branch and Residual module. Unlike traditional algorithms, the model can learn from training data instead of selecting parameters manually such as noise level. Application of this proposed method to the OCT images shows a more than 22 dB signal-to-noise ratio improvement in speckle noise reduction with minimal structure blurring. The proposed method provides strong generalization ability and can process noisy other types of OCT images without retraining. It outperforms other filtering methods in suppressing speckle noises and revealing subtle features.

## 1. INTRODUCTION

Optical coherence tomography (OCT) generates cross-sectional imaging of biological tissue in micron resolution [1]. Speckle noise in OCT limits the visual effect in contrast and impair the clinical diagnosis. Finding an efficient and effective speckle denoising algorithm is a major concern for its theoretical meaning and practice value. A number of image processing methods have been used to despeckle the OCT images, such as median filtering [2], wavelet-based filtering that employs nonlinear thresholds [3], and anisotropic diffusion filtering[4]-[6]. Generally, the outcome can suppress the speckles in areas that are homogeneous. But the noisy images are often blurred or over smoothed resulting in losses in details in nonhomogeneous areas such as structural edges or lines. Kafieh *et al.* proposed an OCT denoising method based on dictionary learning. The atoms in the dictionary can represent the clear image and the noise with less regular structures are removed [7]. But the duration of learning a dictionary is long. Recently, Zhang *et al.* proposed and tested residual learning of deep Convolutional Neural Network (CNN) for natural image denoising, but have not expanded to speckle noise in clinical OCT images [8]. It is a pity that [8] have not used the model of residual. The Residual module makes the proposed model easier to optimize and decreases the training duration. Secondly, the proposed ResNet-based denoising methods in OCT images can handle denoising with unknown noise level or unknown noise type. Thirdly, the network can process input of arbitrary size and generate correspondingly-sized output with efficient learning. Last but not least, based on patch processing instead of entire image, the model can select representative patches that helps to decrease training loss and provides generalization ability for different types of OCT images.

Compared to existing methods, the proposed method provides fast processing speed and has lots of room for time reduction.

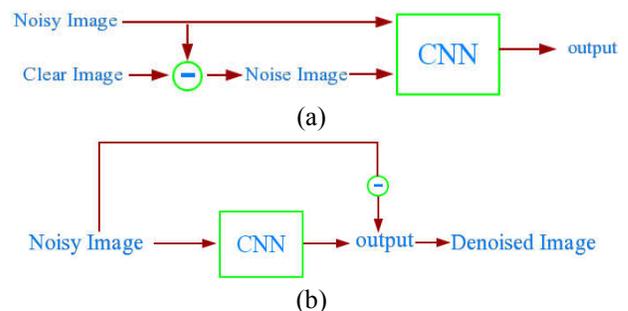

**Fig.1**: (a) Train stage of model (b)Test stage of model

## 2. METHOD

we empirically design an end-to-end fashion network using some design strategy. In the training stage, the whole model learns from the training data and map the noisy image to noise image by minimizing the MSE. In test stage, the network takes the noisy image as input, and output the corresponding noise image. The denoised result is obtained by subtracting the noise image from the noisy one, as shown in Fig.1.

A number of applications have been designed with deep neural network (DNN). For example, CNN has achieved unparalleled success in image classification as the convolution kernel in network can learn diverse features of images. Also non-linearity relation between outcome and input data is learned in optimization stage when the training data flows in models and parameters of convolution kernel are revised. In the following we introduce Residual module and provide details on network architectures as well as the training process and data augmentation.

**2.1 Residual module**

He et al. first proposed the residual module in 2015 [9]. We use residual module in the proposed method, which involves a pairwise addition between input images $x$ and $F(x)$ where $F$ is a functional transformation that contains twice convolutional operation. Residual module is easier to optimize and helps to decrease training duration as it alleviates the degradation problem such as vanishing gradients in model training. The proposed method can gain performance from Residual module with increased depth.

**2.2 The ResNet-based denoising framework**

Let $CBN_{k,s}$ denotes Convolution-BatchNorm-ReLU module with $k$ filters of size $s \times s$, $Res_{k,s}$ denotes a Residual module, as described in 2.1, and $Branch_{k,s}$ denotes a Branch module which is pairwise addition between $CBN_{k,s}(x)$ and $C_{k,s}(C_{k,s}(x))$ where $C_{k,s}(x)$ denotes convolution.

The proposed ResNet-based model can be written as:
$CBN_{64,3} \rightarrow CBN_{64,3} \rightarrow Branch_{64,3} \rightarrow Res_{64,3} \rightarrow Res_{64,3}$
$\rightarrow Res_{64,3} \rightarrow CBN_{64,3} \rightarrow CBN_{64,3}$

As shown in Fig.2, we construct 3 different convolution layer structures as follows to sum up their advantages. (i)Convolution-BatchNorm-ReLU module: filters are used to generate relatively low-level features rather than the input image itself. This module also appeared in last two layers. (ii) Branch module: filters are used to mix diverse features from different branches with different depth. This also increases the width of network, leading to improved performance. (iii) Residual module: as described in 2.1.

Two operations are employed to improve the performance: batch normalization [10] and ReLU activation function [11]. The former improves the training efficiency via reducing the statistical differences between training samples and the latter keeps the sparsity of convolution kernel via restricting the result of convolution.

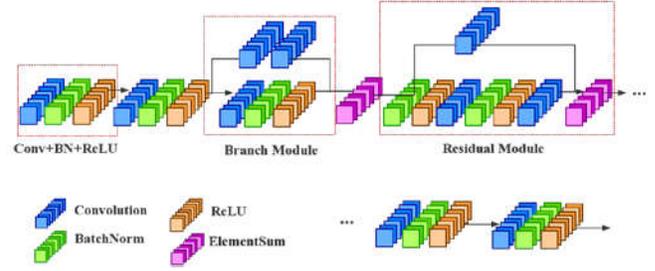

**Fig.2**: Model structure

**2.3 Training**

The size of input and output is $128 \times 128$ and is kept fixed in the model. The depth of the proposed network is 12, which is enough for extracting advanced features while easy to train. Conventionally, the number of convolution kernels in each layer should be more in the middle layers and less in the beginning and end layers, but we set this number to 64 in every layer constrained by actual memory resources. Let $I_C$ denotes the input clear image, and $I_N$ denotes noisy image, $N = (I_N - I_C)$ denotes the noise to be removed and $T(x)$ denotes the mapping transform the network does. We adopt the loss function

$$l(\Theta) = \frac{1}{N \cdot M} \sum_{i=1}^{N} \sum_{j=1}^{M} (T_{i,j}(I_N;\Theta) - (I_{Ni,j} - I_{Ci,j}))^2 \quad (1)$$

to minimize the MSE loss between the output of model $T(I_N)$ and the noise $N$.

**2.4 Training Data Preparation**

For OCT image denoising, there's no ground truth image readily available. We use Bscan averaging to obtain images almost free of speckle noise and use them as training data. M macula-centered 3-D OCT volumes are obtained from the same normal eye. One volume is randomly picked as the target image. For each Bscan in this volume, N nearby Bscans from each of the rest M-1 volumes are registered to it using affine transformation. From the $N(M-1)$ registered images, L images with the highest structural similarity index (SSIM) [12] scores are selected and averaged together with the target Bscan. Hence, the target Bscans and the corresponding averaged results are used as the noisy and denoised images of training set. For this paper, we set M=20, N= 7, and L=10. The Bscans are flattened with respect to the retina bottom, to reduce the size of ROI.

**3. EXPERIMENTAL RESULTS**

The multiple OCT volumes used to generate training data are acquired with the Topcon Atlantis DRI-1 SS-OCT scanner(Topcaon, Tokyo, Japan), with $992 \times 512 \times 256$ (height×width×Bscans) voxels covering a $6 \times 6 mm^2$ macula-centered area. Among the 256 Bscans with ground truth, 246 Bscans are randomly selected as training and validation set. 10-fold cross validation is carried out to observe early stop time. The rest 10 Bscans form the test set, for which the performance indices, such as peak signal-to-noise ratio (PSNR) as structural similarity index (SSIM), can be calculated. To show the method is universally applicable to other types of retinal OCT images. We also test the method on images acquired from different subjects, from other scanners (Topcon OCT-1000 and Zeiss 4000) or other scanning protocols (Topcon Atlantis DRI-1 under wide-view mode, covering $9 \times 6 mm^2$ with both macula and optical nerve head areas). Some of the images show retinal pathologies.

Some overlapping samples are also used in method. Finally, we get 19959 images patches. We set the number of training epochs as 100.

Table I Comparison of performance indices

| Algorithms | Center Bscan | | Peripheral Bscan | |
| --- | --- | --- | --- | --- |
| | PSNR(dB) | SSIM | PSNR(dB) | SSIM |
| Noisy image | 12.74 | -- | 12.53 | -- |
| Median Filter | 31.70 | 0.32 | 31.41 | 0.34 |
| NLM | 32.42 | 0.34 | 32.11 | 0.31 |
| BM3D | 33.43 | 0.35 | 32.80 | 0.39 |
| Proposed | 34.83 | 0.52 | 33.84 | 0.54 |

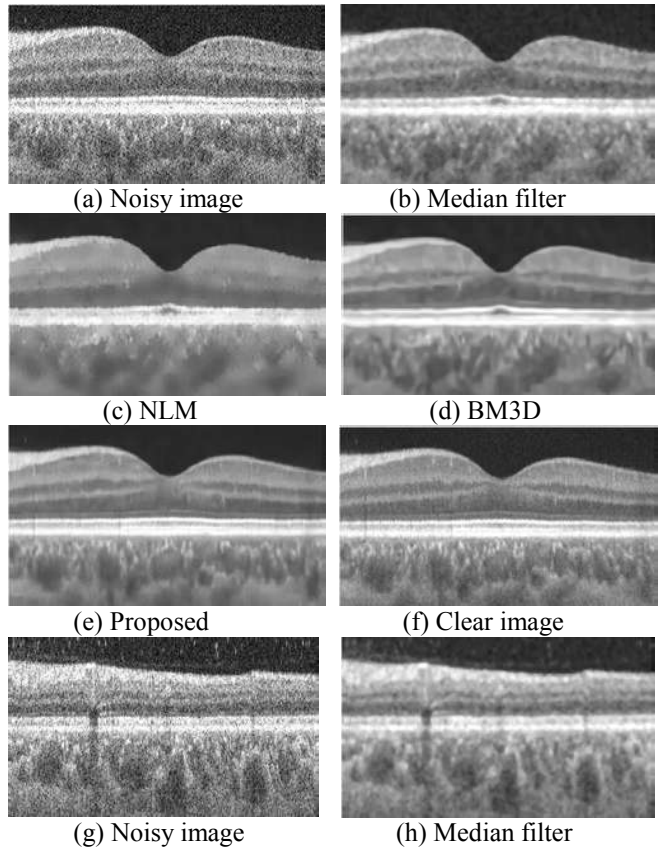

(a) Noisy image (b) Median filter
(c) NLM (d) BM3D
(e) Proposed (f) Clear image
(g) Noisy image (h) Median filter

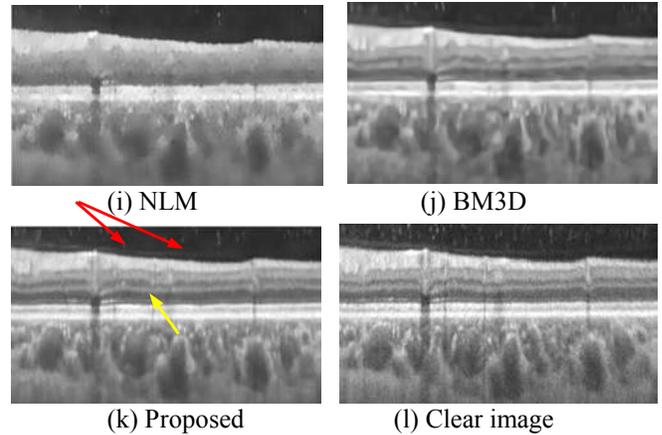

(i) NLM (j) BM3D
(k) Proposed (l) Clear image

**Fig.3** Denoising results using median filer, NLM, BM3D and the proposed method. (a) -(f) from the center Bscan, (g)- (l) the peripheral Bscan.

The results of one center Bscan and one peripheral Bscan from the test set are shown in Fig.3, compared with three other denoising algorithms: median filter, non-local means (NLM) [13], block-matching and 3D filtering (BM3D) [14]. The PSNR and SSIM indices are listed in Table I. The indices are calculated in the ROI that contains the whole retina. In the methods for comparison, the parameters are empirically optimized to give highest PSNR. Computing the PSNR between noisy image and clear image, the proposed method has improved PSNR 20 dB in average.

In Fig.3a and Fig.3g the noisy images have layered structure and are degraded by strong speckle noise. Fig.3b and Fig.3h show that median filter has little effect on removing strong speckle noise and a lot of structural information is lost. Fig.3c and Fig.3i show that NLM algorithm almost removes all noise in background but also removes a lot of important structural details. Fig.3d and Fig.3j show that BM3D keeps the retinal structural details relatively well but the edges are distorted a bit. Fig.3e and Fig.3k show that the proposed method almost keeps all retinal structural details while remove noise well. Especially in Fig.3k, the subtle structures are kept and enhanced well, including the structures inside the vitreous body (red arrows) and the external limiting membrane (yellow arrow).

The denoising results for other types of retinal OCT images is shown in Fig. 4. The proposed method achieves similar speckle reduction performance, showing its generalization ability or adaptability.

Table II Comparison of average running time

| Methods | Median Filter | NLM | BM3D | Proposed |
| --- | --- | --- | --- | --- |
| Average Time(s) | 0.23 | 418.90 | 4.21 | 0.23 |

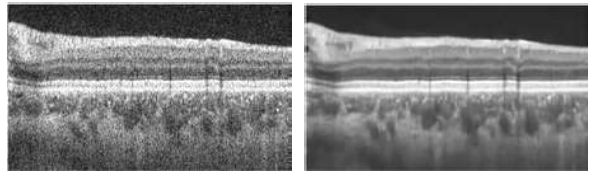
(a)Normal subject, Topcon Atlantis DRI-1, 992×512

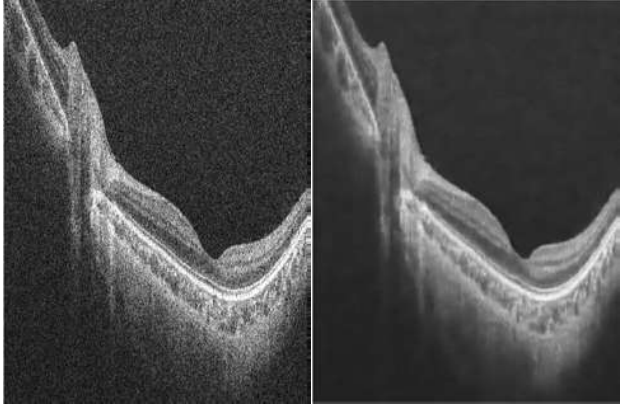
(b)Wide-view, Topcon Atlantis DRI-1, 992×512

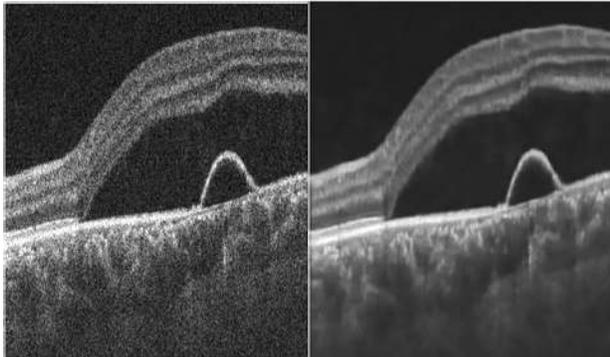
(c) Pathological subject, Topcon Atlantis DRI-1, 992×512

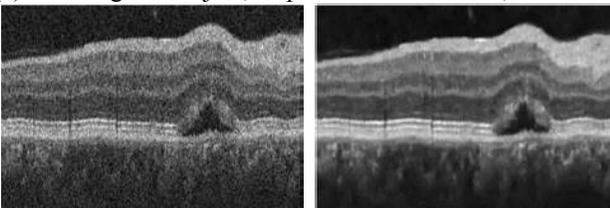
(d) Pathological subject, Zeiss OCT-4000, 1024×512

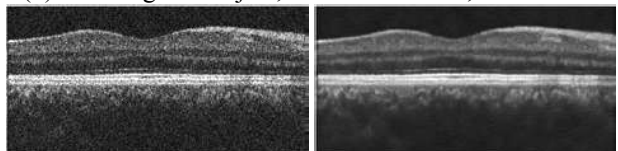
(e) Normal subject, Topcon OCT-1000, 480×512

**Fig.4** Denoising results of other testing images. Column 1 is noisy image and Column 2 is corresponding denoised image

## 4. CONCLUSIONS

In this work, we proposed a ResNet-based universal method to denoise the retinal OCT images which are affected by speckle noise. The following conclusion is reached.

The following conclusions are reached. First, the proposed method yields a relevantly high PSNR and SSIM, indicating that the denoising results are close to the ground truth image obtained by Bscan averaging. Second, the proposed method also has good visual effect, alleviating the noise and keeping the structures at the same time. Third, the processing time of proposed method is short and can almost meet the requirement of clinical diagnosis. Moreover, using more processors or matrix operation optimization, the processing time can be further reduced to reach real-time performance.